%% file: main.tex
\documentclass[11pt]{article}

\PassOptionsToPackage{compress}{natbib}

\usepackage[]{acl}

\usepackage[utf8]{inputenc} 
\usepackage{times}
\usepackage{latexsym}
\usepackage[T1]{fontenc}
\usepackage[utf8]{inputenc}

\usepackage{hyperref}       
\usepackage{url}            
\usepackage{booktabs}       
\usepackage{amsfonts}       
\usepackage{nicefrac}       
\usepackage{microtype}      
\usepackage{xcolor}         
\usepackage{xspace}
\usepackage{amsmath}
\usepackage{amssymb}
\usepackage{graphicx}
\usepackage{comment}

\newcommand{\OpenAIOOne}{\texttt{o1-preview}\xspace}
\newcommand{\Scheherazade}{\emph{Scheherazade}\xspace}
\newcommand{\GSMek}{\texttt{GSM8K}\xspace}
\definecolor{yoshigreen}{HTML}{339933}

\renewenvironment{quote}
  {\list{}{\rightmargin=0.1cm \leftmargin=0.1cm} 
   \item\relax}
  {\endlist}

\title{\Scheherazade: Evaluating Chain-of-Thought Math Reasoning in LLMs with Chain-of-Problems}

\author{
 Stephen Miner$^{*}$ 
 \quad \textbf{Yoshiki Takashima}$^{*}$ 
  \quad \textbf{Simeng Han}$^{*}$  
 \quad \textbf{Sam Kouteili}$^{*}$ \\
 \quad \textbf{Ferhat Erata}$^{*}$
 \quad \textbf{Ruzica Piskac}$^{*}$ 
 \quad \textbf{Scott J Shapiro}$^{*}$ \\
  $^*$Yale University
 }

\begin{document}

\maketitle

\begin{abstract}
  \input{section/abstract.tex}
\end{abstract}

\input{section/intro.tex}
\input{section/related_work}
\input{section/technique.tex}
\input{section/eval.tex}
\input{section/error_analysis}
\input{section/conclusion.tex}
\input{section/limitations.tex}

\newpage


\input{references.bbl}
\input{section/appendix.tex}

\end{document}

%% file: section/abstract.tex
%
Benchmarks are critical for measuring Large Language Model (LLM)
reasoning capabilities. Some benchmarks have even become the de facto indicator of such capabilities. However, as LLM reasoning capabilities improve,
existing widely-used benchmarks such as \GSMek marginally encapsulate model
reasoning differentials - most state-of-the-art models
for example achieve over 94\% accuracy on the \GSMek dataset \cite{gsm8k_res}. While constructing harder benchmarks is possible, their creation is often manual,
expensive, and unscalable. As such, we present {\Scheherazade}, an
automated approach to produce large quantities of challenging mathematical reasoning
benchmarks by \emph{logically chaining} a small starting set of problems. We propose two
different chaining methods, forward chaining and backward chaining,
which include randomized branching techniques to generate complex reasoning problems. We apply Scheherazade on \GSMek to create
\textit{\GSMek-Scheherazade} and evaluate 3 frontier LLMs and OpenAI's
\OpenAIOOne on it. We show that while other frontier models' performance
declines precipitously at only a few questions chained, our
evaluation suggests \OpenAIOOne's performance persists, with the flagship OpenAI model the only one to perform better at backward reasoning. Our data and code are available at https://github.com/YoshikiTakashima/scheherazade-code-data. 

%

%% file: section/intro.tex
%
%
\section{Introduction}
  
\begin{quote}
\vspace{-3pt}
\textit{"No problem can be solved from the same level of consciousness that created it."} - Albert Einstein
\vspace{-3pt}
\end{quote}

Benchmarks serve as a critical tool for evaluating the reasoning abilities of competing LLMs, providing a standardized framework to compare their performance. These benchmarks span a wide range of difficulties, from simple grade-school math problems to challenging math olympiads. By allowing for consistent evaluation across models that are often opaque, proprietary, or both, benchmarks establish a common ground for comparison. Moreover, they play a pivotal role in quantifying LLM development and claims about their reasoning
capabilities, allowing researchers and developers to identify strengths, limitations, and areas for improvement ~\citep{openai2024gpt4technicalreport,
  dubey2024llama3herdmodels, gemmateam2024gemmaopenmodelsbased}.

The reasoning capabilities of LLMs have advanced to the point where performance on many existing mathematical benchmarks has converged to near-perfection \citep{openai2024o1, openai2024gpt4technicalreport, dubey2024llama3herdmodels,gemmateam2024gemmaopenmodelsbased}. This convergence undermines one of the primary purposes of such benchmarks - to differentiate the reasoning abilities of various models. Additionally, the widespread use of existing benchmarks for training and fine-tuning has led to significant data-contamination issues
\citep{zhang2024carefulexaminationlargelanguage,
  matton2024leakagecodegenerationevaluation}. \GSMek in particular has
been rendered less insightful as multiple advanced LLMs have exceeded 94\%
accuracy and competitive performance has been widely achieved \citep{openai2024o1, openai2024gpt4technicalreport,
  dubey2024llama3herdmodels, gemmateam2024gemmaopenmodelsbased, gsm8k, hendrycks2021measuring}.
Despite the rapid consumption and depreciation of benchmarks, novel,
high-quality benchmark sets are limited, and generating new data often
involves costly manual labeling. Synthetic benchmark creation
methods have been proposed, but their scope is limited. Existing
approaches shuffle
sentences~\citep{chen2024premiseordermattersreasoning}, leverage
templates~\cite{zhang2024training, synthesize_math_data} and mutate
constants~\cite{pmlr-v202-gao23f}, limiting the complexity and
diversity of the generated benchmarks.

We introduce \Scheherazade, a 
technique for logically chaining
multiple existing benchmarks together to create larger, more complex benchmark
problems. These problems are designed to test mathematical and logical Chain-of-Thought~(CoT)
reasoning abilities of models. 
To illustrate, consider the statement, "If it rains, I will wear a raincoat." Now, if we modify the statement, for example, to "If 2+3 = 5 and it rains, I will wear a raincoat," we, as humans, can immediately see that this statement is equivalent to the previous one, but the very process of parsing and discarding irrelevant statements requires reasoning. We can add more statements to create a chain of expressions, and find that LLMs struggle to discern meaning as these chains grow. In this paper, we show that such chains are a great way to evaluate LLM reasoning capacity.

Our approach connects benchmarks together using conditional branching. We call this process "chaining benchmarks together" or "benchmark chaining". We chain benchmarks in such a way that the 
necessary information to solve each question in the chain is derived by solving other questions in the chain. We propose two methods of chaining, \emph{forward chaining} and \emph{backward chaining}. In
\emph{forward chaining}, problems are linked using implication such
that the resulting chained problem can be solved sequentially in the order it is written. In contrast, with \emph{backward chaining}, problems
earlier in the chain require contextual information from statements
\emph{later} in the chain. While both methods are logically equivalent, backward chaining introduces an added complexity, as it forces the model to reason in reverse at each step. This unique challenge is analyzed further in the paper, and makes a significant impact on model evaluation results. Our tool leverages conditional branches and randomness to ensure the LLM cannot simply memorize the format. Both chaining techniques are flexible, generalizing to chains of
any length and ordering of their component problems.

We benchmark the mathematical reasoning abilities of four frontier
models - OpenAI o1\cite{openai2024o1}, GPT-4o\cite{openai4o}, Meta Llama 3.1 70B\cite{dubey2024llama3herdmodels}, and Anthropic Claude
3.5 Sonnet \cite{claude3_5} - using a benchmark set created by applying \Scheherazade to
\GSMek. Running the models on our benchmark shows that, despite high reported scores
on original \GSMek problems, performance rapidly declines as the length of the chain increases. Additionally, our analysis of OpenAI's \OpenAIOOne showed it outperforms current
frontier models with longer chains, showing a much more gradual decline in accuracy as the chain length increases. Interestingly, both GPT-4o and \OpenAIOOne show a gradual decline in accuracy chain length increases for forward chaining. However, only \OpenAIOOne follows a similar trend for backward chaining, its accuracy gradually declining with chain length whereas GPT-4o has a steep fall off in accuracy as backward chaining length increases, similar to the other models we evaluated. 
%

%% file: section/related_work.tex
\section{Related Work}

\subsection{Chain of Thought Prompting}

Chain-of-Thought (CoT) prompting has emerged as a powerful technique to enhance the reasoning capabilities of large language models (LLMs) by guiding them to generate intermediate reasoning steps rather than producing direct answers. This approach has proven effective for tasks involving arithmetic, commonsense, and logical reasoning. Recent follow-up work has further advanced CoT prompting through various innovations. For instance, "Self-Consistency" \citep{wang2023selfconsistency} improves CoT by sampling multiple reasoning paths and selecting the most consistent outcome to enhance robustness. Other methods, like "Tree-of-Thoughts" \citep{yao2023tree} explore branching reasoning trajectories to tackle more complex decision-making tasks. These developments illustrate the evolving landscape of CoT prompting and highlight its potential to address increasingly sophisticated reasoning challenges in LLMs.

\subsection{Synthetic math reasoning benchmark creation}

Evaluating mathematical reasoning capabilities in large language models (LLMs) has become increasingly critical as model performance approaches saturation on existing benchmarks. One widely used benchmark, GSM8K, has seen models achieve over 94\% accuracy, undermining its ability to differentiate between advanced LLMs \citep{openai2024gpt4technicalreport, dubey2024llama3herdmodels}. To address this, alternative benchmarks have been proposed to better assess model reasoning, such as the MATH dataset, which presents higher-level competition math problems \citep{hendrycks2021measuring}. However, these datasets often require costly manual curation, making large-scale, automated benchmark generation essential.

Multiple synthetic math reasoning benchmark creation
methods have been recently proposed. \citet{chen2024premiseordermattersreasoning} shuffles the the order of premises in questions of GSM8k, underscoring the need for more sophisticated problem constructions. \cite{zhang2024training, synthesize_math_data} leverage synthetic templates to create. \citet{shi2023largelanguagemodelseasily, anonymous2024gsmsymbolic} add irrelevant context to GSM8k questions and show that it can greatly distract LLMs from the correct reasoning path. \citep{pmlr-v202-gao23f} replace each number in GSM8k questions with a random integer of up to 7 digits and cause chain-of-thought prompting to fail catastrophically. Despite the effectiveness of these methods in testing LLM reasoning capabilities in adversarial scenarios, the complexity and diversity of the generated benchmarks are inherently limited. 

The Scheherazade framework builds on these efforts by introducing problem chaining techniques, specifically forward and backward chaining. Unlike previous methods that rely on isolated problem mutations, chaining creates nested dependencies between problems, significantly increasing the reasoning complexity.  \citet{hosseini2024not} also evaluates LLM reasoning capabilities by chaining math reasoning problems, but they only explore forward chaininig of two problems while we create a more diverse and complex setting.  Scheherazade’s automated benchmark generation aligns with broader trends in synthetic data augmentation and scalable evaluation methods \citep{ye2024physicslanguagemodels21, havrilla2024surveyingeffectsqualitydiversity, liu2024best}. By chaining problems, it offers a novel way to evaluate logical reasoning in LLMs, contributing to the ongoing effort to develop robust benchmarks that can keep pace with rapidly advancing model capabilities.

%% file: section/technique.tex
\section{Approach}



\begin{figure*}[h] \centering
\fbox{\includegraphics[width=14cm]{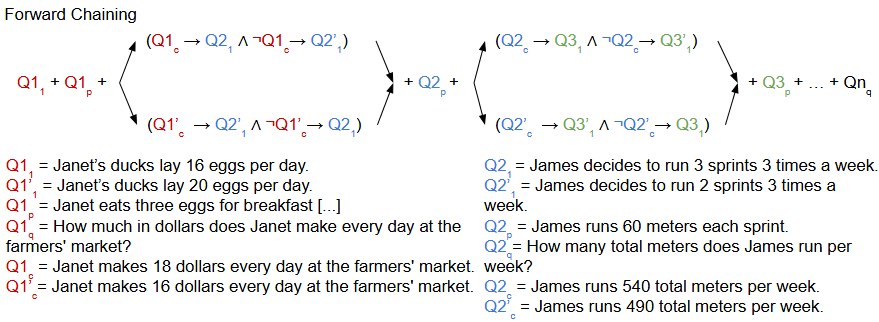}}
  \caption{Forward chaining generalization and example.}
  \label{fig:1}
\end{figure*}

\subsection{Forward Chaining}
We propose two approaches for constructing $n$-length chains of \GSMek problems, where $n$ represents the number of problems in the chain. These problems are
chained together to form a single, composite problem. By introducing branching paths and integrating randomness, we ensure that models cannot simply memorize a sequence of branches to follow, thereby requiring reasoning. The first technique,  \textit{forward chaining}, links problems in a sequential manner, where each problem can be solved in the order it appears. The second
technique, \textit{backward chaining}, reverses this structure: solving any problem in the chain requires information from a subsequent problem. This dependency on future context makes backward chaining uniquely challenging, as the evaluation results show. Both of these techniques can generalize to any chain
length, and support arbitrary ordering of their component problems, meaning that any permutation of problems is allowed.

To formally explain how we
chain problems, we first introduce some notation. For a benchmark problem $P$,
let $Q_1$ be the first logical premise of $Q$. For example, suppose 
\begin{align*}
     Q = &\text{"Alice has 3 apples. Bob has 2 apples. Charlie}  \\ 
     &\text{has 4 apples. How many apples do Alice, Bob,} \\
     &\text{and Charlie have in total?"} 
\end{align*}

then $Q_1 = \text{"Alice has 3 apples."}$ $Q_p$
denotes the remainder of the problem excluding the question; in our example, $Q_p
= \text{"Bob has 2 apples. Charlie has 4 apples"}$. We let $Q_q$ denote the question statement
asking for the solution to the problem. In our example, $Q_q =\text{"How
many apples do Alice and Bob have in total?"}$. $Q_c$ denotes the
conclusion of the problem, written in natural language. For example,
$Q_c = \text{"Alice and Bob have 3 apples in total."}$ We additionally
use $Q_c'$ and $Q_1'$ to refer to a wrong conclusion and
an alternate first premise respectively. Each question in the benchmark comes with a wrong conclusion and an alternate first premise to allow for randomly selecting whether the if conditional statement should be true or false. 

We chain problems together in
two ways, forward chaining and backward chaining. At any point in the
chain, to chain two problems together we create a branching "if then
else" statement. Consider forward chaining with $n=2$ as an example, and let $A$ and $B$ be the two problems. Forward chaining $A$ and $B$
results in one of the two following structures, selected at random:
\begin{align*}
(1) \ \ &A_1 + A_p + (A_c \implies B_1 \land \lnot A_c \implies B_1') \\ 
& + B_p + B_q \\
(2) \ \ &A_1 + A_p + (A_c' \implies B_1' \land \lnot A_c' \implies B_1) \\
&+ B_p + B_q \\
\end{align*}
Here, the $+$ symbol is string concatenation, and $A_c
\implies B_1 \land \lnot A_c \implies B_1'$ is the symbolic equivalent of "If: [$A_c$] is
true, then: [$B_1$] is true, otherwise: [$B_1'$] is true."
Importantly, for any question $Q$, $Q_1'$ has the property that $Q_1'
\nRightarrow Q_c$, meaning if the wrong branch is taken, the corresponding premise resulting from that branch will lead to an incorrect
conclusion. Figure \ref{fig:1} shows how forward chaining generalizes,
and provides two example problems. 
To show how the generalization shown in Figure \ref{fig:1} will look in text form, assume we have a chain of length 2 and the problems $Q1$ and $Q2$ from Figure \ref{fig:1} are the component problems. Then, assume the top path is selected randomly at the first (and in this case only) branching point. Then, the full question will look like the following:
\begin{align*}
&\text{"\textcolor{red}{Janet's ducks lay 16 eggs per day. Janet eats}} \\  
&\text{\textcolor{red}{three eggs for breakfast every morning and bakes}}\\
&\text{\textcolor{red}{muffins for her friends every day with four eggs.}}\\
&\text{\textcolor{red}{Janet sells the remaining eggs at the farmers'}}\\
&\text{\textcolor{red}{market daily for \$2 per fresh duck egg.} If it is}\\
&\text{true that \textcolor{red}{Janet makes 16 dollars every day at}}\\
&\text{\textcolor{red}{the farmers' market}, then \textcolor{blue}{James decides to run}}\\
&\text{\textcolor{blue}{3 sprints 3 times a week.} Otherwise, \textcolor{blue}{James} }\\
&\text{\textcolor{blue}{decides to run 2 sprints 3 times a week. James}}\\
&\text{\textcolor{blue}{runs 60 meters each sprint. How many total}}\\
&\text{\textcolor{blue}{meters does James run per week?}"}
\end{align*}

\subsection{Backward Chaining}
Backward chaining also branches similarly, but unlike forward chaining, requires information from future problems to solve the current problem in the chain. For example, the result of backward chaining problems $A$ and $B$ results in one of the
following, selected at random:
\[(B_c \implies A_1 \land \lnot B_c \implies A_1') + A_p + B_1 + B_p +
A_q \]
\[(B'_c \implies A_1' \land \lnot B_c' \implies A_1) + A_p + B_1 + B_p
+ A_q \] Notice that to get the first premise of $A$, problem $B$ must be solved. However, the premises of problem $B$ do not appear
until after problem $A$. Importantly, notice that in backward chaining
the final question is $A_q$, meaning all intermediate questions must
be solved to solve the final question. As with forward chaining, backward chaining can generalize to any length as follows:
\begin{align*}
&r_1 + Q1_p + r_2 + \dots + Q1_q. 
\end{align*}
Where 
\begin{align*}
r_1 \in \{&(Q2_c \implies Q1_1 \land \lnot Q2_c \implies Q1_1'), \\
&(Q2_c' \implies Q1_1' \land \lnot Q2_c' \implies Q1_1)\} \\
r_2 \in \{&(Q3_c \implies Q2_1 \land \lnot Q3_c \implies Q2_1'), \\ 
&(Q3_c' \implies Q2_1' \land \lnot Q3_c' \implies Q2_1)\} \\
\vdots
\end{align*}
Where $r_1, r_2$, etc. are chosen at random,  and ending with the question, $Q1_q$.

This generalization shows that as chain length increases, the
reasoning required to solve the problem becomes increasingly
nested. That is, information from $Q2$ is required to resolve the
first premise of $Q1$, with information from $Q3$ necessary to resolve $Q2$, and so on. Note that the final question asked is $Q1_q$, the first problem in the chain. This is because obtaining $Q1_1$ requires solving every problem from $Q2$ to $Qn$. In other words, the question asked always corresponds to the problem with the longest preceding chain of reasoning required to solve it. 
We can use the same problems $Q1$ and $Q2$ from before to illustrate an example of backward chaining. Assume that the choice $(Q2'_c \implies Q1'_1 \land \lnot Q2_c' \implies Q1_1)$ was selected at random. Then, the example will look like that following when chained backward:
\begin{align*}
&\text{"If it is true that \textcolor{blue}{James runs 490 total meters}}\\
&\text{\textcolor{blue}{per week}, then \textcolor{red}{Janet's ducks lay 20 eggs per day.}}\\
&\text{Otherwise, \textcolor{red}{Janet's ducks lay 16 eggs per day.}}\\
&\text{\textcolor{red}{Janet eats three eggs for breakfast every morning}}\\
&\text{\textcolor{red}{and bakes muffins for her friends every day with}} \\
&\text{\textcolor{red}{four eggs. Janet sells the remaining eggs at the}}\\
&\text{\textcolor{red}{farmers' market daily for \$2 per fresh duck egg.}}\\
&\text{\textcolor{blue}{James decides to run 3 sprints 3 times a week.}} \\ 
&\text{\textcolor{blue}{James runs 60 meters each sprint.} \textcolor{red}{How much in}}\\
&\text{\textcolor{red}{dollars does Janet make every day at the farmers'}}\\
&\text{\textcolor{red}{market?}"}
\end{align*}
Notice how in order to solve the backward chaining problem, which asks a question about the first problem in the chain, the second problem in the chain needs to be solved. In this case, it has to be determined whether or not James runs 490 total meters before being able to determine a piece of information crucial to evaluating how much Janet makes every day at the farmers' market. 

\subsection{Generation Power}
Besides generating more complex problems, our technique allows for the generation of a large number of problems given a small starting set. Starting with $N$ unchained benchmarks and given chain length $l$, there are $P(N,l) = \frac{N!}{(N-r)!}$ ways to order $l$ questions. This means that starting with $N$ benchmarks, there are $\sum_{l=2}^{N} \frac{N!}{(N-r)!}$ ways to select questions to be chained. Then, with the randomness introduced between each subsequent question in the chain, and taking into account there are two techniques, the total number of benchmark problems the Scheherazade technique can generate given $N$ starting benchmark problems is: 
\[2 \cdot \sum_{l=2}^{N} \frac{2^{l-1} \cdot N!}{(N-l)!}.\]
For example, with just 50 benchmark problems to start with ($N=50$), our technique can generate approximately $5.64 \cdot 10^{79}$ different problems.


%% file: section/eval.tex
%
%
\section{Evaluation}
\label{section:eval}
We use \Scheherazade to extend \GSMek and create \GSMek-Scheherazade. We evaluate each of GPT-4o (Aug. 6th 2024), \OpenAIOOne, Anthropic Claude 3.5 Sonnet, and
Meta Llama 3.1 70BGP on chains of length 2 through 10 with both forward chaining and backward chaining. For all models except \OpenAIOOne, we evaluate each model on 1000 randomly generated forward and backward chaining problems for each length, for a total of 18,000 benchmark cases per model (9000 forward, 9000 backward, at lengths 2-10). This amounts to a total of 54,000 benchmark cases across the three models. Due to budget and API constraints, \OpenAIOOne is evaluated on 200 problems per chain length per technique for a total of 3600 problems.

\begin{table*}[]
\centering
\caption{Raw accuracy numbers up to length
  10. Despite near-perfect
  performance by frontier models at length 1 (original \GSMek
  problems), the performance rapidly declines for most models.}
\resizebox{\textwidth}{!}{
\begin{tabular}{r|llllllllll}
\toprule
Length &1 & 2 & 3 & 4 & 5 & 6 & 7 & 8 & 9 & 10 \\
\midrule
  \textbf{Forward} \\
Claude 3.5 &     0.986 &      0.280 &      0.302 &      0.274 &      0.240 &      0.236 &      0.197 &      0.177 &      0.173 &      0.156 \\
gpt-4o &     0.971 &      0.920 &      0.870 &      0.795 &      0.785 &      0.730 &      0.710 &      0.700 &      0.550 &      0.500 \\
Llama3.1 70B &     0.971 &      0.268 &      0.187 &      0.124 &      0.067 &      0.044 &      0.015 &      0.011 &      0.007 &      0.005 \\
\OpenAIOOne & 1.000 & 0.960 & 0.970 & 0.935 & 0.910 & 0.860 & 0.890 & 0.845 & 0.865 & 0.815 \\
\midrule
  \textbf{Backward} \\
Claude 3.5 &     0.986 &      0.879 &      0.599 &      0.319 &      0.179 &      0.102 &      0.074 &      0.056 &      0.045 &      0.032 \\
gpt-4o &     0.971 &   0.970 &      0.720 &      0.550 &      0.350 &      0.260 &      0.140 &      0.120 &      0.110 &      0.055 \\
Llama3.1 70B &     0.971 &      0.477 &      0.265 &      0.113 &      0.064 &      0.035 &      0.015 &      0.014 &      0.002 &      0.001 \\
\OpenAIOOne &     1.000 &      0.980 &      0.965 &      0.960 &      0.945 &      0.935 &      0.920 &      0.915 & 0.945 &  0.925 \\
\bottomrule
\end{tabular}
}
\vspace{0.2pt}
\label{tbl:raw}
\end{table*}

\begin{figure*}[t]
  \centering
    \fbox{\includegraphics[width=0.75\textwidth]{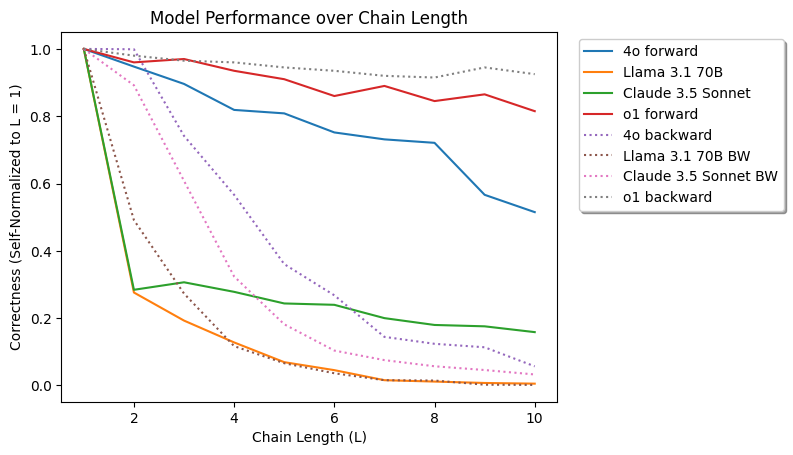}}
    \caption{Accuracy of LLMs declines when the chains become
      longer. With the exception of \OpenAIOOne, LLMs find backward
      chains harder than forward chains at longer lengths.}
  \label{fig:eval}
\end{figure*}

\paragraph{Raw Accuracy} Table~\ref{tbl:raw} presents the results of our evaluation, showing the raw accuracy (scaled between 0 and 1) for each chain length. The results reveal a consistent decline in accuracy for all models as chain length increases, for both forward chaining and backward chaining. This supports the notion that logical question chaining creates more challenging reasoning problems, allowing us to better differentiate the reasoning capabilities of the different models. At longer lengths, every model except \OpenAIOOne fails at backward reasoning, unable to reason about almost every question by chain length 10. Forward chaining presents similar results with the exception of GPT-4o, which declines more slowly, reaching a 0.5 raw correctness accuracy by length 10. It is interesting to note that while Claude 3.5 performs marginally better than GPT-4o at a chain length of one, from chain lengths of 2 onward, its backward reasoning results are worse, and its forward counterparts are drastically worse. These observations raise questions about potential overfitting to prominent benchmarks such as \GSMek.

\paragraph{Normalized Accuracy} Fig.~\ref{fig:eval} displays the performance normalized to accuracy on
problems of length 1 (standard \GSMek problems). The
horizontal axis denotes the number of chained questions, while the
vertical axis represents normalized accuracy. Normalized to original \GSMek performance as per Fig.~\ref{fig:eval}, we see sharp accuracy declines exhibited by every model except \OpenAIOOne for backward chaining. By chains of length 2-3, normalized backward chaining accuracy plummets for every model other than \OpenAIOOne, which maintains relatively strong accuracy over longer chain lengths. Normalized accuracy also declines significantly for forward chaining for every model except \OpenAIOOne and GPT-4o, with the former exhibiting similar results across chain lengths and the latter's accuracy declining more gradually than its counterparts. It is noteworthy that GPT-4o has the highest disparity between forward chaining accuracy and backward chaining accuracy - the former remains relatively high across all lengths, while the latter sharply declines in a similar, though slightly more gradual, manner to other models. 

Notably, for shorter lengths, all models present better backward reasoning numbers with shorter questions than forward counterparts, but exhibit a sharper accuracy decline - by length 6, the latter generally overtakes the former. \OpenAIOOne remains an outlier here, with backward chaining maintaining higher accuracy than forward chaining. These findings spark an interesting conversation in model reasoning patterns. If we consider the fundamental nature of large language models as agents that \emph{predict the next token}, it makes sense that models perform better at forward reasoning, as they are sequentially selecting next tokens after scanning the input problem. It's interesting then, that \OpenAIOOne, the newest model, is able to overcome this limitation and exhibit even higher accuracy on backward reasoning than it does on forward reasoning. Though the reasons for this cannot be entirely known due to the opaque, proprietary nature of \OpenAIOOne and other OpenAI models, these results could spark an interesting discussion about what could have changed to elicit such drastically different behavior on backward reasoning. This topic warrants further investigationn, and more experiments need to be done to narrow down the difference between \OpenAIOOne and its predecessors as it related to backwards reasoning.


%% file: section/error_analysis.tex
\section{Error Analysis}
\label{section:error_analysis}

\begin{figure*}[t] \centering
\fbox{\includegraphics[width=0.75\textwidth]{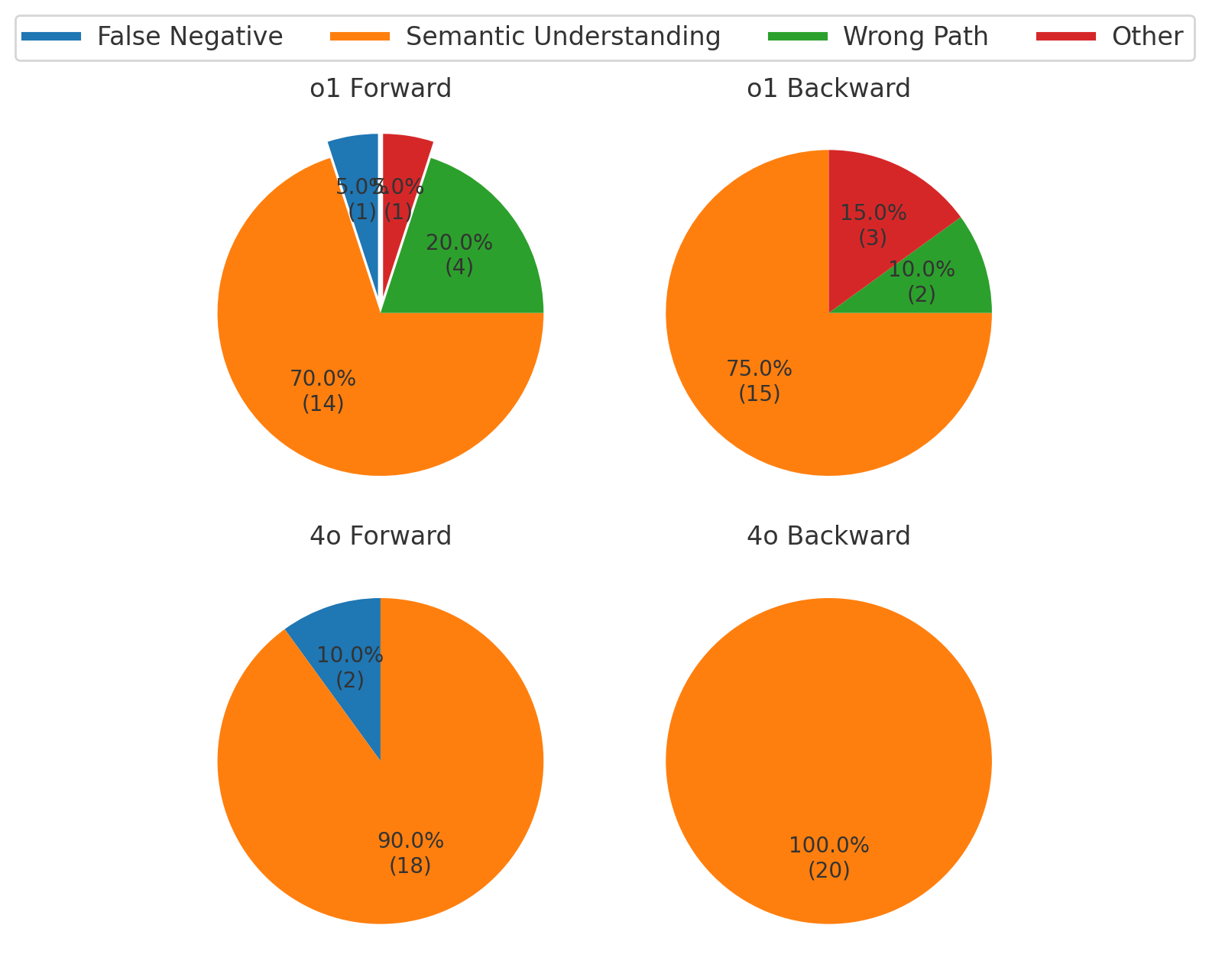}}
\caption{Pie charts showing the error categorization of errors made by
  \OpenAIOOne and GPT-4o, both for forward and backward
  chaining.}
  \label{fig:errors}
\end{figure*}

In addition to overall accuracy results, we randomly sample 40 incorrect benchmark answers produced by both GPT-4o and \OpenAIOOne: 20 forward chaining and 20
backward chaining errors. We manually analyze these errors and find that we can
categorize them into four categories: semantic misunderstanding
error, wrong path error, false negative, and other. We present these findings in Fig.~\ref{fig:errors}. These error categories differ from those in \cite{CoT} as they were either no longer present or not suitably expressive here - there were for example no instances of "calculation errors" in our findings as LLMs are now able to run computations, and missing step errors are now encompassed in a broader wrong path error category. 

\paragraph{Semantic Understanding Errors} Semantic understanding errors, which make up the largest portion of errors, occur when the model produces an answer indicating a misunderstanding of the question asked. This can be any semantic understanding problem, including, for example, not understanding something within an individual problem or a misunderstanding of
the condition statements. An example of the former can be found in a benchmark asking for Stephen's total grocery costs - here, it interprets a 25\% tip on the 40 dollar bill as stating the original bill was $40/1.25$, as opposed to $40 \times 1.25$. The latter is a more generalizable error - here the LLM often starts by assuming the question statement and works backwards. For example, in a question about the number of ripe oranges, the conditional chain constructed by {\Scheherazade} is "If it is true there are 17 ripe oranges, then suppose ..." - both \OpenAIOOne and GPT-4o occasionally run into the error of assuming there are 17 ripe oranges, and working backward to fit the premises to the conclusion, showing a fundamental misunderstanding of the question.  This latter semantic understanding error type is particularly interesting, as it highlights a fundamental flaw in a model's ability to reason about logical control flows. In trying to over-fit to the numbers present in the question, the models are not able to semantically parse what the question is asking.

\paragraph{Wrong Path Errors} The second kind of errors are wrong path errors. Here, one of three things happens: either the LLM solved an intermediate problem correctly and then followed the wrong conditional branch, the LLM skipped a step entirely, or the LLM hallucinated a new alternative path that did not exist in the question. These hallucinations refer to instances where the model was able to reason through a question in the chain correctly, but then in trying to fit to the presented number in the conditional chain, presents what it calls an \verb|Alternative Exploration|, where the model tweaks some numbers in its proof to fit what it has calculated is the desired answer.

\paragraph{False Negatives} A false negative occurs when our evaluation system fails to properly extract the answers from the LLM's output, despite the LLM answering correctly. This appears, for example, when the LLM proves that the conditional branch is false instead of finding the correct answer to the question and using that to refute the conditional branch. This is because our extraction system looks for the answers to each intermediate question, and by refuting the conditional statement the correct answer to the question sometimes is not explicitly stated. Finally, "other", the red pieces of the pie charts, captured errors that did not fit into the other error categories. The number of these errors was relatively small, so we believe our error categories effectively captured the errors. 

\paragraph{Infinite Loops} Although it was not in our random sample of errors, we noticed one interesting behavior that only GPT-4o exhibited. On some questions, GPT-4o would end up in the equivalent of an infinite loop. The model would perform an incorrect calculation, and instead of trying something else, it just repeated the calculation over and over until hitting the output limit and throwing an error back to our local machine. We know this only occurred with GPT-4o because only GPT-4o produced outputs that were occasionally significantly longer than all other outputs (around 100,000 characters long).

Although Fig.~\ref{fig:errors} only displays a randomized subset of the total errors made by the models, we find that interesting trends appear. Our error analysis only highlights semantic understanding errors for GPT-4o, while showing a wider array of errors for \OpenAIOOne. One hypothesis for this is that \OpenAIOOne is more capable of understanding logically complex questions. If a semantic error occurs, GPT-4o may not even get to generate content that contains other types of errors. In contrast, because \OpenAIOOne was able to avoid many semantic errors, other error types were allowed to emerge. More work is needed to determine if this is indeed the case. Nevertheless, we believe the error analysis and evaluation performed offer valuable insight into these models' reasoning capabilities.




%% file: section/conclusion.tex
%
\section{Conclusion and Future Work}

Benchmarks are the foundation upon which the current language model
ecosystem stands. Their rapidly eroding value is a cause for concern.
We presented \Scheherazade, a novel framework for extending reasoning benchmark complexity, and the generation of large amounts of benchmarks given a small starting set. By constructing composite problems through forward and backward chaining techniques, we extend \GSMek to create \textit{\GSMek-Scheherazade}, though any set of starting benchmarks can be used. We apply these new benchmarks on existing frontier models and find significant performance changes, creating notable evaluation of model reasoning capabilities that was no longer achieved by the native \GSMek dataset. In doing so, we also garnered a better understanding of model forward versus backward reasoning capabilities, and gained perspective on the different ways \OpenAIOOne and GPT-4o struggled with certain benchmark cases through our error analysis.


The results of running \GSMek-Scheherazade on frontier models suggests
several avenues for future work.
First, applying \Scheherazade to math reasoning benchmarks other
than \GSMek such as MATH (Mathematical Problem Solving Dataset) would aid in slowing their depreciation.
Second, statements combining logical processes other than if-then-else should be
explored. Combining problems with conjunctions or
disjunctions in addition to implications may further nuance our understanding of model reasoning patterns. With numerical benchmarks, numerical operators such as taking sums of solutions would also be relevant.
Third, combining \Scheherazade with more fine-grained re-orderings of the
questions remains an interesting avenue to explore. While we presented purely backward and forward chaining, hybrid combinations may allow us to figure out the scope of, in particular, \OpenAIOOne's reasoning abilities, which maintains strong reasoning results. More fine-grained mutations within questions rather than at the entire question level could also prove interesting. The combination of within-question mutations with our chaining techniques could serve to expand \Scheherazade's benchmark generation abilities and allow for further increases in difficulty as-needed. 


%% file: section/limitations.tex
\section{Limitations}
While our \Scheherazade framework offers a novel approach to evaluating LLM reasoning capabilities, it is not without its limitations. One notable challenge lies in the method used to extract answers from the model outputs. This approach occasionally produces false negatives, as the model's response may be correct but formatted in an unexpected way. In our error analysis, we found that 3.75\% of the errors were due to such false negatives, potentially underestimating model performance in some cases.

Another limitation pertains to the requirements for the base set of benchmarks used in chaining. Although  the \Scheherazade technique can create a large number of benchmarks from a small starting set, for the starting set of benchmarks to be suitable, they must adhere to specific constraints. They cannot use pronouns, and the names of people and variables across benchmarks must be unique to avoid ambiguity. Additionally, an incorrect conclusion must be created for each problem in the starting set of benchmarks to enable the random selection of conditional branches within our techniques.  While some aspects of this preprocessing can be automated, certain datasets may require some amount of manual cleanup to meet these requirements. This additional step, though manageable, introduces some overhead when applying \Scheherazade.

%% file: section/appendix.tex
\label{section:example_problems}
\appendix
\section{Forward Chaining Example Problems}
\subsection{Length 2}
Emily has 4 kids named Amy, Jackson, Corey, and Jason. Amy is 5 years older than Jackson and 2 years younger than Corey. Jason is 10. Jason is 1 year younger than Corey. If it is true that Jackson is exactly 5 years old, then the following is true: [ Joseph takes care of 12 dogs.] Otherwise, the following is true: [Joseph takes care of 10 dogs.] Each dog that Joseph takes care of takes .5 hours a day to walk and take care of their business. How many hours per week does Joseph spend taking care of dogs?\\

\textbf{solutions}: [4, 35]
\subsection{Length 4}
A raspberry bush has 6 clusters of 20 fruit each and 67 individual fruit scattered across the bush. If it is true that There are exactly 177 raspberries on the raspberry bush in total, then the following is true: [ Mishka bought 2 pairs of shorts, 3 pairs of pants, and 3 pairs of shoes.] Otherwise, the following is true: [Mishka bought 3 pairs of shorts, 3 pairs of pants, and 3 pairs of shoes.] One pair of shorts that Mishka bought costs \$16.50. One pair of pants that Mishka bought costs \$22.50 and one pair of shoes costs \$42. If it is true that Mishka spent exactly 247 dollars total on all of the clothing items combined, then the following is true: [ The great dragon, Perg, sat high atop mount Farbo, breathing fire upon anything within a distance of 1150 feet.] Otherwise, the following is true: [The great dragon, Perg, sat high atop mount Farbo, breathing fire upon anything within a distance of 1000 feet.] Polly could throw the gold javelin, the only known weapon that could sleigh the dragon, for a distance of 400 feet, well within the reach of the dragon's flames.  But when Polly held the sapphire gemstone, she could throw the javelin three times farther than when not holding the gemstone. Polly is holding the gemstone. If it is true that Polly could stand at most 100 feet outside the reach of the dragon's flames and still hit the dragon with the gold javelin, then the following is true: [ Toulouse has three times as many sheep as Charleston.] Otherwise, the following is true: [Toulouse has twice as many sheep as Charleston.] Charleston has 4 times as many sheep as Seattle. Seattle has 20 sheep. How many sheep do Toulouse, Charleston, and Seattle have together? \\

\textbf{solutions}: [187, 243, 200, 260]

\subsection{Length 6}
Jim spends 2 hours watching TV and then decides to go to bed and reads for half as long. Jim watches TV and read 3 times a week. If it is true that Jim spends exactly 28 hours on TV and reading in four weeks, then the following is true: [ Gloria is shoe shopping when she comes across a pair of boots that fit her shoe budget. Gloria has to choose between the boots she came across and two pairs of high heels that together cost seven dollars less than the boots.] Otherwise, the following is true: [Gloria is shoe shopping when she comes across a pair of boots that fit her shoe budget. Gloria has to choose between the boots she came across and two pairs of high heels that together cost five dollars less than the boots.] One pair of heels Gloria is looking at costs \$33. The other pair of heels that Gloria is looking at costs twice as much as the \$33 pair. If it is true that The boots that Gloria came across are exactly 98 dollars, then the following is true: [ The great dragon, Perg, sat high atop mount Farbo, breathing fire upon anything within a distance of 1150 feet.] Otherwise, the following is true: [The great dragon, Perg, sat high atop mount Farbo, breathing fire upon anything within a distance of 1000 feet.] Polly could throw the gold javelin, the only known weapon that could sleigh the dragon, for a distance of 400 feet, well within the reach of the dragon's flames.  But when Polly held the sapphire gemstone, she could throw the javelin three times farther than when not holding the gemstone. Polly is holding the gemstone. If it is true that Polly could stand at most 200 feet outside the reach of the dragon's flames and still hit the dragon with the gold javelin, then the following is true: [Dan plants 3 rose bushes.] Otherwise, the following is true: [Dan plants 2 rose bushes.] Each rose bush Dan plants has 25 roses. Each rose on Dan's rose bushes has 8 thorns. If it is true that There are exactly 600 thorns in total on all of Dan's roses, then the following is true: [Meredith is a freelance blogger who writes about health topics and submits to clients each day as her permanent job. A blog article takes Meredith an average of 4 hours to research and write about.] Otherwise, the following is true: [Meredith is a freelance blogger who writes about health topics and submits to clients each day as her permanent job. A blog article takes Meredith an average of 5 hours to research and write about.] Last week, Meredith wrote 5 articles on Monday and  2/5 times more articles on Tuesday than on Monday. On Wednesday, Meredith wrote twice the number of articles she wrote on Tuesday. If it is true that Meredith spent exactly 106 hours writing articles in three days, then the following is true: [ Joseph takes care of 12 dogs.] Otherwise, the following is true: [Joseph takes care of 10 dogs.] Each dog that Joseph takes care of takes .5 hours a day to walk and take care of their business. How many hours per week does Joseph spend taking care of dogs?\\

\textbf{solutions}: [36, 104, 200, 600, 104, 35]

\section{Backward Chaining Example Problems}

\subsection{Length 2}
If it is true that Christina will spend exactly 22 dollars on gift bags, then the following is true: [Jan has four times the number of pets as Marcia.] Otherwise, the following is true: [Jan has three times the number of pets as Marcia.] Marcia has two more pets than Cindy. Cindy has four pets. Christina is planning a birthday party and needs .75 gift bags per invited guest, because 1/4 of attendees don't show up. Christina invited 16 friends. Gift bags cost \$2 each. How many total pets do Jan, Marcia, and Cindy have total?\\

\textbf{solutions}: [24, 28]

\subsection{Length 4}
If it is true that There were exactly 615 gems in the buried treasure chest, then the following is true: [James decides to run 2 sprints 3 times a week.] Otherwise, the following is true: [James decides to run 3 sprints 3 times a week.] James runs 60 meters each sprint. If it is true that Billy sold exactly 9 DVDs on Tuesday, then the following is true: [A treasure hunter found a buried treasure chest filled with gems. There were 200 diamonds, 35 fewer rubies than diamonds, and twice the number of emeralds than the rubies in the buried treasure chest.] Otherwise, the following is true: [A treasure hunter found a buried treasure chest filled with gems. There were 175 diamonds, 35 fewer rubies than diamonds, and twice the number of emeralds than the rubies in the buried treasure chest.]  If it is true that Joshua still has exactly 3 lego sets, then the following is true: [Billy sells DVDs. He has 8 customers on Tuesday. Billy first 3 customers buy two DVDs each.] Otherwise, the following is true: [Billy sells DVDs. Billy has 8 customers on Tuesday. Billy's first 3 customers buy one DVD each.] Billy's next 2 customers buy 2 DVDs each.  Billy's last 3 customers don't buy any DVDs. Joshua plans to sell all his toys and use the money to buy video games. Joshua has 13 lego sets and he sells them for \$15 each. Joshua ends up buying 8 video games for \$20 each and has \$5 left. How many total meters does James run per week? \\

\textbf{solutions}: [2, 7, 595, 540]

\subsection{Length 6}
If it is true that There are exactly 187 raspberries on the raspberry bush in total, then the following is true: [Lloyd has an egg farm. Lloyd's chickens produce 252 eggs per day and he sells them for \$2 per dozen.] Otherwise, the following is true: [Lloyd has an egg farm. Lloyd's chickens produce 273 eggs per day and he sells them for \$2 per dozen.]  If it is true that The average square footage of a level of Luke's sandcastle is exactly 80 square feet, then the following is true: [A raspberry bush has 5 clusters of 20 fruit each and 67 individual fruit scattered across the bush.] Otherwise, the following is true: [A raspberry bush has 6 clusters of 20 fruit each and 67 individual fruit scattered across the bush.]  If it is true that There were exactly 615 gems in the buried treasure chest, then the following is true: [Luke is spending time at the beach building sandcastles. Luke eventually notices that each level of a sandcastle will have double the square footage as the level below it.] Otherwise, the following is true: [Luke is spending time at the beach building sandcastles. Luke eventually notices that each level of a sandcastle will have half the square footage as the level below it.] Luke makes a 4 leveled sandcastle. The top level of the sandcastle that Luke makes has a square footage of 16. If it is true that Christina will spend exactly 24 dollars on gift bags, then the following is true: [A treasure hunter found a buried treasure chest filled with gems. There were 175 diamonds, 35 fewer rubies than diamonds, and twice the number of emeralds than the rubies in the buried treasure chest.] Otherwise, the following is true: [A treasure hunter found a buried treasure chest filled with gems. There were 200 diamonds, 35 fewer rubies than diamonds, and twice the number of emeralds than the rubies in the buried treasure chest.]  If it is true that Wendi needs to give her chickens exactly 20 cups of feed in the final meal of the day, then the following is true: [Christina is planning a birthday party and needs .75 gift bags per invited guest, because 1/4 of attendees don't show up. Christina invited 16 friends.] Otherwise, the following is true: [Christina is planning a birthday party and needs .75 gift bags per invited guest, because 1/4 of attendees don't show up. Christina invited 20 friends.] Gift bags cost \$2 each. Every day, Wendi feeds each of her chickens three cups of mixed chicken feed, containing seeds, mealworms and vegetables to help keep them healthy. Wendi gives her chickens their feed in three separate meals. In the morning, Wendi gives her flock of chickens 15 cups of feed.  In the afternoon, Wendi gives her chickens another 25 cups of feed. The size of Wendi's flock is 20 chickens. How much does Lloyd make on eggs per week?\\

\textbf{solutions}: [20, 24, 595, 60, 187, 294]